\documentclass[letterpaper, 10 pt, conference]{ieeeconf}  

\IEEEoverridecommandlockouts                              

\usepackage{graphicx}
\usepackage{siunitx}
\usepackage{amsmath}
\usepackage{amsfonts}
\usepackage{mathtools}
\usepackage{amsthm}
\usepackage{color}
\let\labelindent\relax 
\usepackage{enumitem}

\theoremstyle{plain}
\newtheorem{assumption}{Assumption}

\newcommand{\fig}{Fig.~}

\newcommand{\set}[1]{\boldsymbol{\mathbf{#1}}}

\title{\LARGE \bf
Decentralization of Multiagent Policies\\by Learning What to Communicate
}

\author{James Paulos$^*$, Steven W. Chen$^*$, Daigo Shishika, and Vijay Kumar%
\thanks{We gratefully acknowledge the support of ARL grant ARL DCIST CRA W911NF-17-2-0181, ONR grant N00014-07-1-0829, and ARO grant W911NF-13-1-0350. This work was supported in part by the Semiconductor Research Corporation (SRC) and DARPA. We additionally thank NVIDIA for generously providing support through the NVAIL program. }%
\thanks{The authors are with the GRASP Lab at the University of Pennsylvania, Philadelphia, USA. {\{\tt\small {jpaulos, chenste, shishika, kumar\}@seas.upenn.edu}}}%
\thanks{$^*$\textit{These authors are co-first authors.}}
}

\begin{document}

\maketitle
\thispagestyle{empty}
\pagestyle{empty}

\begin{abstract}

Effective communication is required for teams of robots to solve sophisticated collaborative tasks.
In practice it is typical for both the encoding and semantics of communication to be manually defined by an expert; this is true regardless of whether the behaviors themselves are bespoke, optimization based, or learned.
We present an agent architecture and training methodology using neural networks to learn task-oriented communication semantics based on the example of a communication-unaware expert policy.
A perimeter defense game illustrates the system's ability to handle dynamically changing numbers of agents and its graceful degradation in performance as communication constraints are tightened or the expert's observability assumptions are broken.

\end{abstract}

\section{INTRODUCTION}

Teams of cooperative robots hold unique advantages over single actors across many task domains.
Examples include mapping~\cite{saeedi_multiplerobot_2016, howard_multirobot_2006}, search \cite{liu_multirobot_2016,kantor_distributed_2006}, precision agriculture \cite{barrientos_aerial_2011,kazmi_adaptive_2011}, and logistics and delivery \cite{dandrea_revolution_2012}.
There are, however, fundamental challenges to designing effective multi-robot systems.
The first is the high dimensionality of a team's joint state space, which makes the planning problem difficult even in a full information setting.
The second is that in realistic settings the information to support those decisions is spread throughout the team.
As a result, today's deployed systems often require a central authority and/or effectively unrestricted communication, both of which can lead to brittle solutions with limited scalability.

This work envisions a team of robots which work collectively as peers towards a common goal.
Teammates can be dynamically added or removed, disrupting no explicit hierarchy or rigid agent specialization.
Our methodology is to train agents as a team to mimic the actions of an expert policy.
During training the expert policy may be centralized, full information, poorly scaling, communication-naive; during deployment the agents will make due with local information and well defined communications constraints.

The key to task success is effective communication, but it is far from obvious how to manually engineer it.
Information can be irrelevant, redundant, or salient but non-actionable.
Messages themselves may wish to convey summaries of state, intended actions, or commands to other agents.
We co-train agents' communication and decision processes as layered neural networks, where a small number of activations represent network links between agents in the team.
In this way we learn communications that support the task.

Evolutionary methods can be used to co-evolve agents represented by neural networks~\cite{miikkulainen_multiagent_2012}. Beyond these methods, there is a rich history of multi-agent reinforcement learning (MARL)~\cite{busoniu_comprehensive_2008}, where a common approach is centralized training of decentralized policies~\cite{kraemer_multiagent_2016,oliehoek_optimal_2008}. Recently, many works have explored multi-agent analogues to single-agent deep reinforcement learning algorithms~\cite{lowe_multiagent_2017,foerster_counterfactual_2018}. These prior works focus on reinforcement learning as a paradigm to train a network, whereas our focus is on structuring the network in a way to decentralize known centralized policies.

Deep reinforcement learning has been applied to to multi-agent problems to explore learned communication protocols.
In \cite{foerster_learning_2016} agents learn to encode meaning in discrete communicative actions (``on/off'') to solve puzzles such as the ``Switch Riddle.'' A common element of these games is a rigid number of participants and strictly serial and sequential communication. Nearer to our vision, \cite{sukhbaatar_learning_2016} considers multi-turn games with agents in a grid world.
Outside of reinforcement learning, \cite{dobbe_fully_2017} views communication as a selective exchange of private state and seeks policies to maximize the mutual information between states and optimal actions.

The primary contribution of this paper is an architecture for training decentralized agent policies based on examples of centralized, expert team play.
This necessarily entails learning communication strategies, and the framework makes explicit that effective communication is a task-oriented property.
Section II formulates our search for a distributed policy and the limitation of when such solutions can be expected to exist.
Section III notes characteristics unique to mobile robot teams that will inform our architecture, which is detailed in Section IV.
We introduce a perimeter defense game in Section V to demonstrate our architecture, evaluate performance in Section VI, and discuss future work in Section VII.

\section{DECENTRALIZATION PROBLEM}
A \textit{multiagent Markov Decision Process (MMDP)}~\cite{boutilier_planning_1996} is a tuple
\begin{equation*}
\begin{aligned}
\mathcal{M} = \langle \mathcal{X}, \set{\alpha}, \{\mathcal{U}_{i}\}_{i \in \set{\alpha}}, \set{f}, q, T \rangle
\end{aligned}
\end{equation*}
where: $\mathcal{X}, \set{\alpha}$ are sets of states and agents; $\mathcal{U}_{i}$ is the set of possible actions for agent $i$; $\set{f}(x, \{u_{i}\}_{i \in \set{\alpha}}): \mathcal{X} \times \{\mathcal{U}_{i}\}_{i \in \set{\alpha}} \rightarrow  \Delta \mathcal{X}$ is a transition rate function; $q(x,\{u_{i}\}_{i \in \set{\alpha}}): \mathcal{X} \times \{\mathcal{U}_{i}\}_{i \in \set{\alpha}} \rightarrow \mathbb{R}$ is a cost rate function; and $T$ is a finite time horizon.

In the \textit{centralized} setting, the objective is to find an optimal centralized policy $\set{\Phi}^{*}_{c}(x):\mathcal{X} \rightarrow \{\mathcal{U}_{i}\}_{i \in \set{\alpha}}$ such that $\forall x \in \mathcal{X}$, $\set{\Phi}^{*}_{c}$ minimizes:
\begin{equation}
\label{eq:general_problem}
\begin{aligned}
& \underset{\set{\Phi}}{\text{min}}
& & J(x) = \int_{t=0}^{T} q(x(t),\set{\Phi}(x(t)))\\
& \text{s.t.}
& & \frac{dx(t)}{dt} = \set{f}(x(t),\set{\Phi}(x(t))), \\
& & & x(0) = x.
\end{aligned}
\end{equation}

In the \textit{distributed setting}, there is a private observation function $y_{i}(x):\mathcal{X} \rightarrow \mathcal{Y}_{i}$, where $\mathcal{Y}_{i}$ is the observation space for agent $i$.
The team observation function is $Y(x): \mathcal{X} \rightarrow \{\mathcal{Y}_{i}\}_{i \in \set{\alpha}}$, which corresponds to the set of private observations $Y(x) = \{y_{i}(x)\}_{i \in \set{\alpha}}$. These observation functions are set by the environment and not the agents.

We make the assumption that the state can be determined using only the current team observation:
\begin{assumption}[Jointly Fully Observable]
    \label{assump:jointly_fully_observable}
    There exists a surjective function $g(\{y_{i}\}_{i \in \set{\alpha}}): \{\mathcal{Y}_{i}\}_{i \in \set{\alpha}} \rightarrow \mathcal{X}$.
\end{assumption}
The MDP under this assumption has been previously called a \textit{decentralized MDP} (as opposed to a \textit{decentralized POMDP}) in other contexts~\cite{amato_decentralized_2013}.
With this assumption, each agent $i$ has a communication function and action function:
\begin{equation*}
\begin{aligned}
\psi_{i}(y_{i}):~& \mathcal{Y}_{i}  \rightarrow C_{i} \\
\mu_{i}(y_{i}, \{\psi_{j}\}_{j \in \set{\alpha}} ):~&  \mathcal{Y}_{i} \times \{C_{j}\}_{j \in \set{\alpha}} \rightarrow \mathcal{U}_{i}
\end{aligned}
\end{equation*}
where $\psi_{i}$ computes the communication an agent broadcasts to other agents based on its private observation $y_{i}$; and $\mu_{i}$ computes the action based on the private observation $y_{i}$ and received communications $\{\psi_{j}\}_{j \in \set{\alpha}}$.

The \textit{distributed policy} $\set{\Phi}_{d}(\{y_{i}\}_{i \in \set{\alpha}}): \{\mathcal{Y}_{i}\}_{i \in \set{\alpha}} \rightarrow \{\mathcal{U}_{i}\}_{i \in \set{\alpha}}$ is the set of predicted actions $\set{\Phi}_{d}(y) = \{u_{i}\}_{i \in \set{\alpha}}$ given the communication and action functions. In this setting, the objective is to find a set of optimal communication and action functions $\{\psi_{i}^{*}\}_{i \in \set{\alpha}}$, $\{\mu_{i}^{*}\}_{i \in \set{\alpha}}$ such that $\forall x \in \mathcal{X}$, the resulting distributed policy $\set{\Phi}^{*}_{d}$ minimizes Eqn.~\eqref{eq:general_problem}.

Frequently, it may be difficult to directly search for optimal distributed policies, but optimal or close to optimal centralized policies are readily available, thus motivating the following problem.

\textbf{Decentralization Problem}: Given a centralized policy $\set{\Phi}_{c}$, the objective of the decentralization problem is to find the set of communication and action functions $\{\psi_{i}\}_{i \in \set{\alpha}}$, $\{\mu_{i}\}_{i \in \set{\alpha}}$ such that $\set{\Phi}_{c}(x) = \set{\Phi}_{d}(\{y_{i}\}_{i \in \set{\alpha}})$. Assumption~\ref{assump:jointly_fully_observable} is a sufficient, but not necessary, condition that the decentralization problem is feasible.

Note that this problem only seeks to find the distributed analogue to a centralized policy, and does not directly deal with optimizing Eqn.~\eqref{eq:general_problem}. However, in most contexts, $\set{\Phi}_{c}$ is found by minimizing Eqn.~\eqref{eq:general_problem}. As a result, the objective function $J(x)$ can be used as a measure of quality for $\set{\Phi}_{d}$.

\section{PROPERTIES}
\label{Sec:Properties}
We are interested in problems concerning \textbf{homogeneous agents} which are defined as having the same communication and action functions $\psi$, $\mu$. In addition, individuals in a team may fail, and it is desirable to have the policy $\set{\Phi}$ be applicable for teams of \textbf{variable size}. The codomain of policy $\set{\Phi}$ is then the set of all sets of arbitrary size of the form $\{\mathcal{U}, \dots, \mathcal{U}\}$ , \textit{i.e.} the power set $2^{\mathcal{U}}$~\cite{zaheer_deep_2017}.

The restriction to homogeneous agents results in a specific invariance property for the policy. Intuitively, swapping two agents with each other should not have an effect on what a third agent does. This intuition is formalized by the concept of \textbf{permutation invariance}~\cite{zaheer_deep_2017}. A function $g:2^{\mathcal{S}} \rightarrow \mathcal{Z}$ acting on sets is permutation invariant if for any permutation $\pi$, $g(\{s_{1}, \dots s_{M}\}) = g(\{s_{\pi(1)}, \dots s_{\pi(M)}\}$.

In addition, at the team level, the set of actions $\{u_{i}\}_{i \in \set{\alpha}}$ should automatically rearrange themselves as the agents swap order. This notion is formalized by the concept of \textbf{permutation equivariance}~\cite{ravanbakhsh_deep_2017, ravanbakhsh_equivariance_2017}. A function $g:\mathcal{S}^{M} \rightarrow \mathcal{Z}^{M}$ is permutation equivariant if for any permutation $\pi$, $g(\pi([s_{1}, \dots s_{M}])) = \pi(g([s_{1}, \dots s_{M}]))$.

In the \textit{MMDP} with homogeneous agents, the following invariances must hold for the action function and policy:
\begin{enumerate}[label={\arabic{enumi}}.,ref={Step \arabic{enumi}},leftmargin=*]
    \item \textit{The action function $\mu$ is \textit{permutation invariant} to the communications $\psi$, and as a result, to the agents $\set{\alpha}$}.
    \item \textit{The policy $\set{\Phi}$ is \textit{permutation equivariant} to the agents $\set{\alpha}$}.
\end{enumerate}

For our specific scenario, our agents observe interchangeable objects in the environment. We would like to impose that the communication function $\psi$ is permutation invariant to these objects, and should also be able to handle variable sizes of these objects.

\section{TRAINING ARCHITECTURE}




Neural networks are suitable for the decentralization problem, and multi-agent systems in general, due to their connectionistic approach to computation.
\textit{When properly designed, a neural network that represents the distributed policy $\set{\Phi}_{d}$ for the team can be shattered into sub-networks to recover individual communication and action functions $\psi$ and $\mu$ for agents}. This insight implies that by operating at the aggregate team level, for example training the network using supervised learning to fit the centralized policy $\set{\Phi}_{c}$, the communication and decision functions will automatically be learned.

\begin{figure*}
    \centering
    \includegraphics[width=7in]{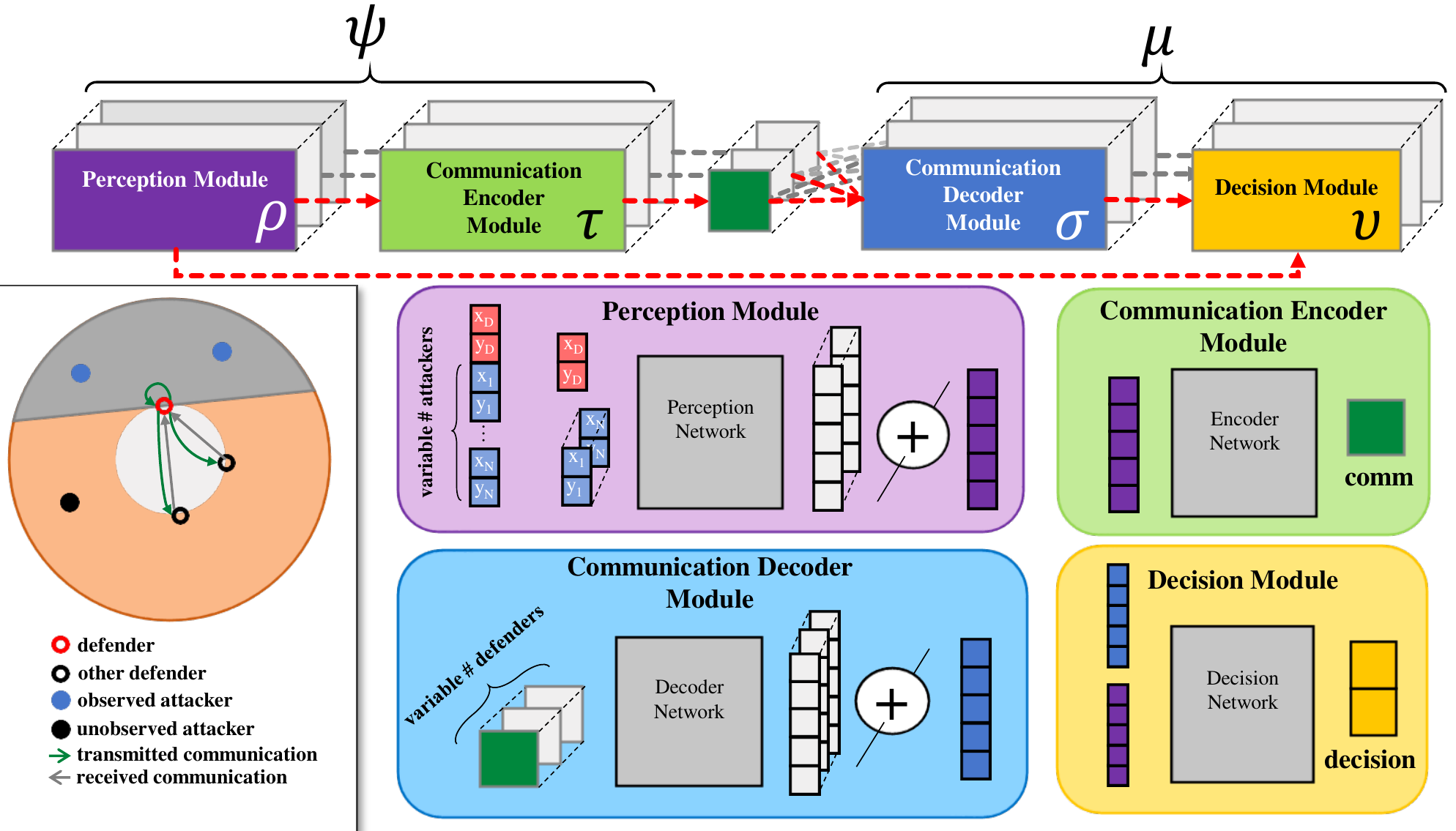}
    \caption{Network Architecture. The communication $\psi$ and decision $\mu$ are functions of the individual agents. The whole diagram represents the entire distributed policy $\set{\Phi}_{d}$ across all agents. }
    \label{fig:network_architectures}
    \vspace{-5mm}
\end{figure*}

Using multi-layer perceptron (\textit{MLP}) and permutation invariant neural network (\textit{PIN}) components, we can impose structure on the network such that it satisfies the properties of Sec.~\ref{Sec:Properties} by construction.
An \textit{MLP} $\phi(s;\theta)$ with $L$ layers is a composition of $L$ affine functions $\lambda_j(s) := W_{j}s + b_{j}$, each except the last one followed by a nonlinear activation function $h$, so that:
\[
\phi(s; \theta) = \lambda_{L} \circ h \circ \lambda_{L-1} \circ \cdots \circ h \circ \lambda_{1}(s),
\]
where $\theta:=\{W_{1:L}, b_{1:L}\}$ are the affine function parameters to be optimized, and $h$ is a fixed function which we choose to be a rectified linear unit (ReLU) or leaky rectified linear unit (Leaky ReLU) \cite{goodfellow_deep_2016}.
The \textit{MLP} is a \textit{fixed-to-fixed} function that takes in a vector of fixed size, and outputs a vector of fixed size.
More structured architectures can be constructed by assigning some of the parameters beforehand, for example setting a specific weight and bias to be 0 in order to remove a connection, or by sharing weights across or between layers.

The \textit{PIN}~\cite{zaheer_deep_2017} is a neural network that takes as input a set of instances and outputs a vector of fixed size. A \textit{PIN} consists of two \textit{MLP}s $\phi$, $\gamma$ arranged in the form $\chi(S) = \phi\left(\sum_{s \in S} \gamma(s) \right)$, where the input is a set $S$. The \textit{PIN} thus destroys the ordering in the set $S$ through the summation operation. In addition, due to weight sharing, it is a \textit{variable-to-fixed} neural network, since the same MLP $\gamma$ is applied to each instance $s \in  S$.

Fig.~\ref{fig:network_architectures} depicts an architecture that obeys the properties highlighted in Sec.~\ref{Sec:Properties}, and can be easily shattered to obtain the communication and action functions $\psi$ and $\mu$ for each agent. The colored blocks on the top of Fig.~\ref{fig:network_architectures} depict the sub-network for a single agent. It consists of $2$ \textit{PIN}s in order to handle $2$ types of permutation invariance: action function $\mu$ is permutation invariant to communications and agents; and communication function $\psi$ is permutation invariant to the objects. Within each \textit{PIN}, there are $2$ \textit{MLP}s, thus our architecture consists of $4$ \textit{modules}, with each module shown to have a different color. These $4$ modules are replicated for each agent in the team to form the distributed policy $\Phi_{d}$, represented by the entire diagram. This replication process enforces both the homogeneity in the agents, as well as the variable size property of the distributed policy.

The composition of these modular blocks leads to natural interpretation for each module. The first module is an MLP $\rho(y): \mathcal{Y} \rightarrow \mathbb{R}^{p}$, called the \textbf{perception module}, and the second module is an MLP $\tau(\rho): \mathbb{R}^{p} \rightarrow C$ called the \textbf{communication encoder module}. The perception module is responsible for perceiving the world and processing it into some meaningful information called a perception feature $\rho$, and the communication encoder module takes $\rho$ and computes the communication to be transmitted.

Usually, we want to impose permutation invariance to certain parts of the observations, for example to semantic objects in the environment, but not other parts such as position or velocity information, i.e. $y = \{\omega_{i}\}_{i \in \set{\Omega}} \times \bar{y}$. As a result, our perception module $\rho$ consists of two sub-network \textit{MLP}s, $\rho_{a}$ and $\rho_{d}$, where $\rho_{a}$ processes the object information we want to be invariant to, and $\rho_{d}$ processes other inputs. The perception feature is then the concatenation of the outputs of these two sub-networks $\rho = [\rho_{d}(\bar{y}), \sum_{i \in \set{\Omega}} \rho_{a}(\omega_{i})]$, and the composition $\tau \circ \rho$ is a \textit{PIN} that computes the communication $\psi(y)$ and is invariant to the objects.

These communication functions $\tau \circ \rho$ are then run on every agent in the team, to get the set of communications $\{\psi_{i}\}_{i \in \set{\alpha}}$. These \textbf{communications} are one of the most fundamental components of the neural network, as they represent the actual messages that will be sent between the agents of the teams. The width of these messages, \textit{i.e.} the size of the message vector, should be small, since transmitting large messages is not desirable. As a result, they represent a bottleneck layer that is a compressed representation of the perceived information.

The third module is an MLP $\sigma(\{\psi_{i}\}_{i \in \set{\alpha}}): 2^{C} \rightarrow \mathbb{R}^{q}$, called the \textbf{communication decoder module} that takes in the set of communications from the team and outputs a decoded feature $\sigma$. The fourth module is an MLP $\nu(\rho \times \sigma): \mathbb{R}^{p+q} \rightarrow \mathcal{U}$, called the \textbf{decision module}. The action function is then computed $\mu(y, \{\psi_{i}\}_{i \in \set{\alpha}}) = \nu \left( \left[ \rho(y), \sum_{i \in \alpha}\sigma(\psi_{i}) \right] \right)$,
and is a \textit{PIN} that is permutation invariant to the communications, and as a result, the agents.

These action functions $\nu \circ \sigma$ are run on every agent in the team to get the set of individual actions $u_{i}$. $\Phi_{d}$ is computed by aggregating these individual actions into a set $\Phi_{d} = \{u_{i}\}_{i \in \set{\alpha}}$. This aggregation procedure ensures that the network is permutation equivariant to the agents, as swapping the order of input of the agents will correspondingly swap the output actions.

The replication of the modules across the agents ensures that they are homogeneous since they each have the same communication and action functions. The usage of the \textit{PIN} architecture allows for the neural network to handle variable sizes of both agents and objects, and also imposes the desired properties of permutation invariance and equivariance. In addition, the \textit{PIN} architecture naturally results in a modular composition with semantic interpretation, and the joining of the $2$ \textit{PIN}s creates a bottleneck layer representing the actual transmitted and received communication messages.

We specified that message width should be small, and this notion can be made more precise by looking at the number of bits needed to represent that message. We can use tools from other areas of deep learning research to express this idea. Training with limited-precision numbers~\cite{gupta_deep_2015,courbariaux_binaryconnect_2015} is an orthogonal area of research in neural networks where the motivation is to use lower precision representations in order to speed up computation and energy efficiency without sacrificing performance. The conversion of float to limited-precision representations is a technique known as quantization. It has a clear application in the context of multi-agent systems, since the quantization technique can be placed on the bottleneck communication layer as a means of controlling message size. We can thus use quantization to make a message small by limiting both the message width size, as well as the number of bits each message can use.

\section{PERIMETER DEFENSE GAME}

\begin{figure}
\centering
\includegraphics[]{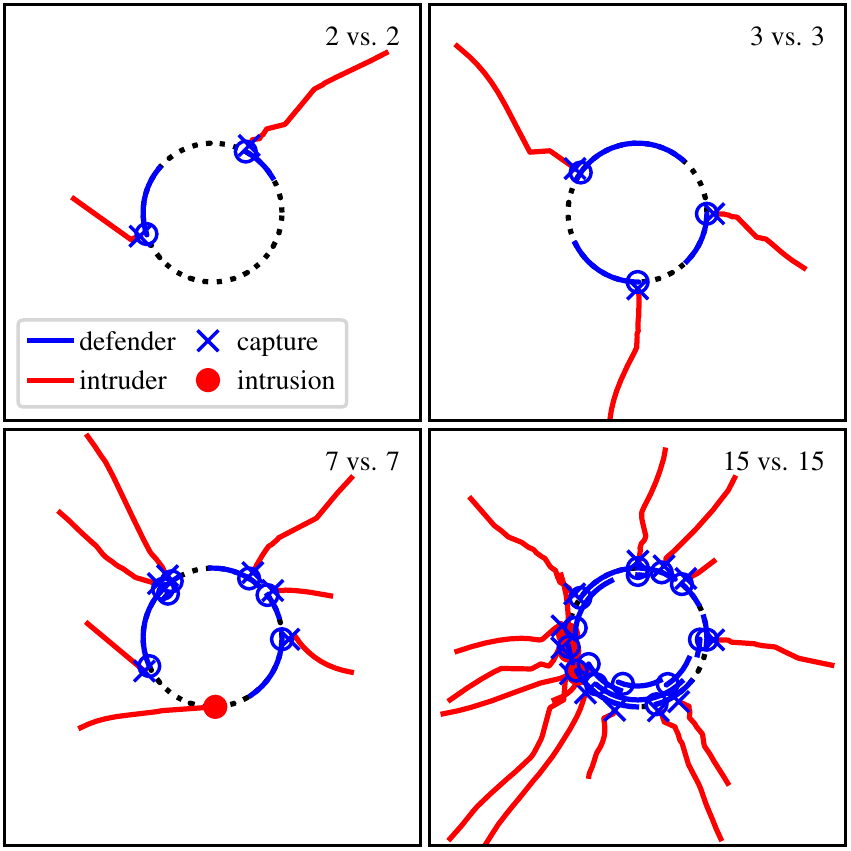}
\caption{Simulated games where defenders execute the learned policy. Overlapping defender paths are offset for clarity.}
\label{fig:sim_path}
\vspace{-5mm}
\end{figure}

We evaluate our architecture in a team perimeter defense game that naturally introduces dynamically varying numbers of teammates as well as varying degrees of private and shared information.
A team of defenders move on the perimeter of a unit circle and seek to capture incoming intruders before the intruders reach the circle interior.
Several simulated games are illustrated in \fig\ref{fig:sim_path}.
All agents move at unit speed, defenders capture intruders by closing within a distance \(\epsilon=0.1\), and both defender and intruder are consumed during capture.

In the controls literature this scenario has been formulated as a full information differential game \cite{shishika_localgame_2018}.
Expert policies for both the defenders and intruders have been introduced based on the concept of maximum matching assignments between defenders and intruders.
These assignment-based strategies are combinatorial in nature and assume centralized information and full observability.
Nevertheless, we use this expert, centralized, communication-unaware policy to train a team of cooperative defenders who share information through learned communication channels.

In our version of this game, each defender knows only its own position and the position of its observed intruders as \(x,y\) coordinates.
Any information about other defenders' locations or unseen intruders' locations will need to come through active communication.
We consider both the situation where defenders see all intruders within a \SI{360}{\degree} field of view (making the joint state strictly observable) and the more difficult case where defenders' observations are limited to an outwardly directed \SI{180}{\degree} field of view as depicted at the left of \fig\ref{fig:network_architectures}. This version of the game violates Assumption~\ref{assump:jointly_fully_observable}, as some attackers may not be seen by any defender.

\section{EVALUATION}
We train the neural network using supervised learning where the the expert $\Phi_{c}$ is the maximum matching algorithm. We randomly sample $10$ million examples from each possible scenario (\textit{e.g.} 5 vs. 3) and query $\Phi_{c}$ for the expert action to generate a dataset.

Each network $\rho_{a}$, $\rho_{d}$ of the perception module has 5 layers of size $1024$, and the perception feature vector is of size $2048$. The remaining modules each have $3$ layers of size $1024$, and the decoded feature is of size $2048$. Each layer is followed by the Leaky ReLU activation function. We train the network in TensorFlow on an NVIDIA DGX-1 using a batch size of $1024$ and a learning rate of $3e^{-4}$ that is dropped by a third every $50,000$ iterations. We randomly sample from the dataset using a manually defined biased sample that weights more difficult scenarios higher. Total training time for one network takes $\sim12$ hours for $200,000$ iterations.

\subsection{Simulation Methodology}

Games are instantiated with equal numbers of defenders and intruders starting from random position.
We simulate games to completion in order to evaluate how learning to replicate the expert's action choices at training time leads to successful online outcomes.
As the games progress, both the size of the team and the number of observed targets change as defenders and intruders are consumed.
The intruders execute an intelligent but uncooperative policy motivated by \cite{shishika_localgame_2018}.

Simulations for several games are depicted in \fig\ref{fig:sim_path}, where the defense team are trained agents with an \SI{180}{\degree} field of view restriction.
From some random starting configurations it is fundamentally impossible for any defense policy to capture all intruders.
For this reason we evaluate the performance of our learned policies by comparing the number of intruders captured to the expert policy.
In the shown 7~vs.~7 game the missed intruder is also missed by the expert policy, and in fact could never be captured.
In the 15~vs.~15 game the trained model slightly underperforms the expert policy.
The figure shows three of the fifteen intruders are missed, but the expert policy misses only one.


\subsection{Visualizing Actions}

\begin{figure}
\centering
\includegraphics[]{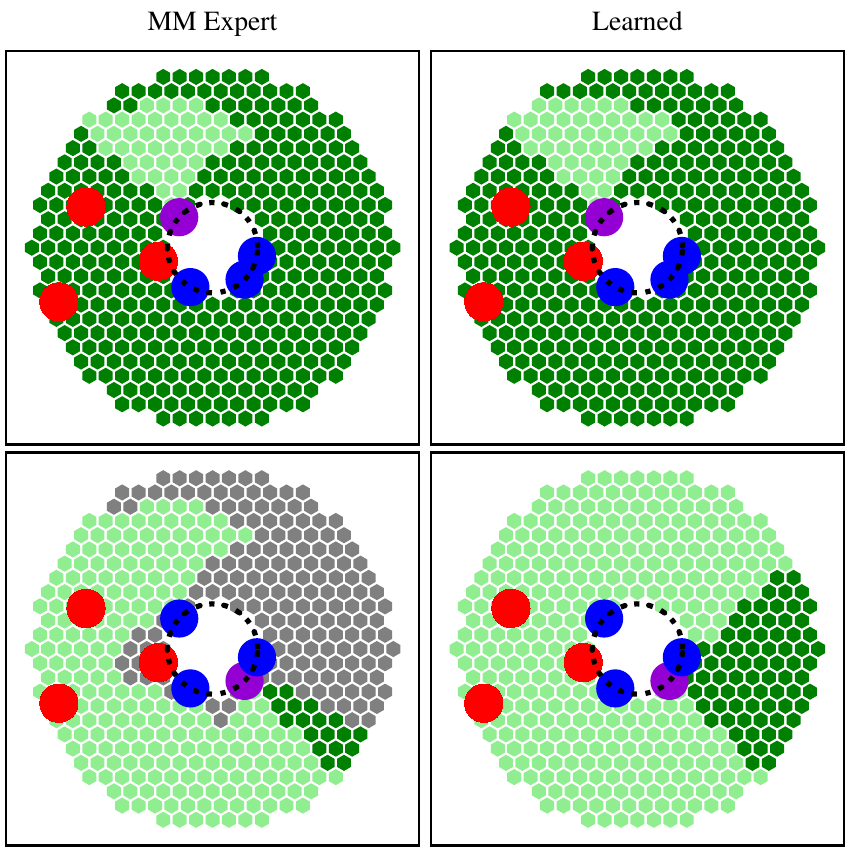}
\caption{Actions of one defender as a function of one intruder's location in a 4~vs.~4 game.  Dark green is move counterclockwise, light green is clockwise, and in grey it does not matter.}
\label{fig:action_regions}
\end{figure}

Each agent faces the decision of choosing one action based on available information about the \((n_d + 2 n_a)\)-dimensional game state\footnote{While not explicit in the representation, the defender positions lie on a 1-dimensional manifold (the circle).}.
A slice of this continuous decision domain is sampled in \fig\ref{fig:action_regions} for a 4~vs.~4 game.
The top left diagram illustrates the appropriate action for the highlighted purple defender as a function of the position of one roaming intruder.
The three other defenders and three other intruders are held fixed in their shown locations.
The defender should move counterclockwise if the final intruder is in a dark green region, and clockwise if it lies in a light green region.
The top left figure shows this defender should usually move left, but can be called upon to move right if an intruder appears in the small light green region.
The bottom left figure illustrates actions for a different defender in the team, and now includes grey regions where the assignment-based expert policy chooses not to employ that defender.

Note that the precise location of an intruder can affect the required defender response even if the intruder is far from the defender.
Intuitively, the appearance of an intruder can trigger a cascade of necessary reassignments that ultimately affect seemingly unrelated defenders.

The right column of \fig\ref{fig:action_regions} shows the defender actions elected by the trained network based on peer communication.
The top figure shows a fairly faithful rendering of the expert policy.
Below we see that the trained network has simplified decision boundaries, because when the expert does not use a defender, the trained networks choice does not matter.

\subsection{Varying Team Size}

\begin{figure}
\centering
\includegraphics[]{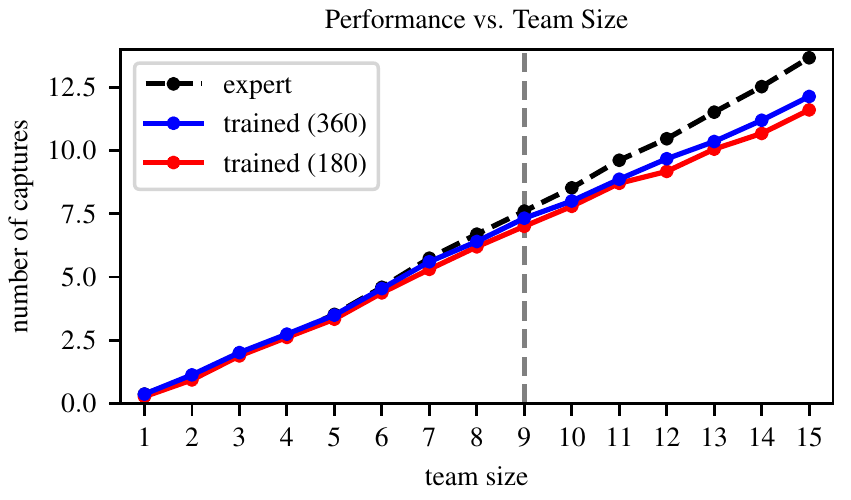}
\caption{Average number of captures in varying size \(n\) vs. \(n\) games.}
\label{fig:score_vs_team_size}
\vspace{-5mm}
\end{figure}

The network was trained using examples of teams of up to size 9; in testing we look at performance in teams of up to size 15.
\fig\ref{fig:score_vs_team_size} depicts the average number of captures in \(n\)~vs.~\(n\) games of increasing size.
We compare the centralized expert policy (using the full game state) to learned policies employing a message width of size one under both \SI{360}{\degree} and \SI{180}{\degree} fields of view constraints.
Team performance closely approximates the expert policy in spite of communications bottlenecks up to the trained team size.
As teams grow much larger than those seen during training, performance begins to lag the expert policy, but the decline in performance is graceful and does not seem to indicate a catastrophic loss of team cohesion and cooperative play.

\subsection{Varying Message Width}

\begin{figure}
\centering
\includegraphics[]{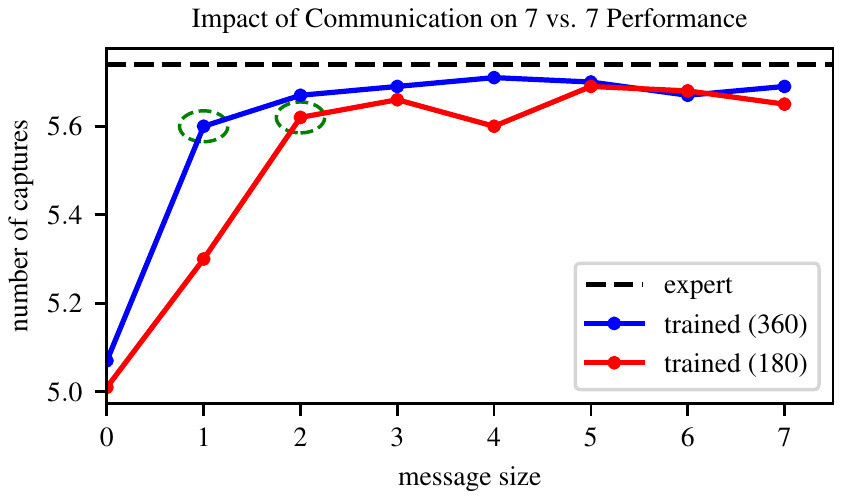}
\vspace{-7mm}
\caption{Score vs. communication width.}
\label{fig:score_vs_comm_size}
\vspace{-5mm}
\end{figure}

We compare the performance of models trained with varying message widths in \fig\ref{fig:score_vs_comm_size}.
Messages are tuples of quantized 8-bit values, and we try constraining the message length anywhere from from length \(0\) (no communication) to length \(7\).
Both trained distributed policies perform poorly with respect to the baseline centralized policy when no communication takes place.

The policy with a full \SI{360}{\degree} field of view of intruders attains near expert-level performance at a message width of 1 (all agents broadcast a single 8-bit value), and increasing the message width does not significantly raise performance.
Each defender need only hear all the other defenders' locations to infer the full joint state, and agents have learned to encode this information into a single scalar value.

An effective message encoding when the field of view is limited to \SI{180}{\degree} is more difficult to engineer.
Each defender is not merely the only teammate to know its own location, it might be the only teammate to see any number of attackers.
In spite of this difficulty, the trained model achieves near expert performance with a message size of only two 8-bit values.
This is suggestive that the agents are not merely summarizing local observations -- they are learning to discern task-oriented information that supports the decision process.

\begin{figure}
\centering
\includegraphics[width=3.4in]{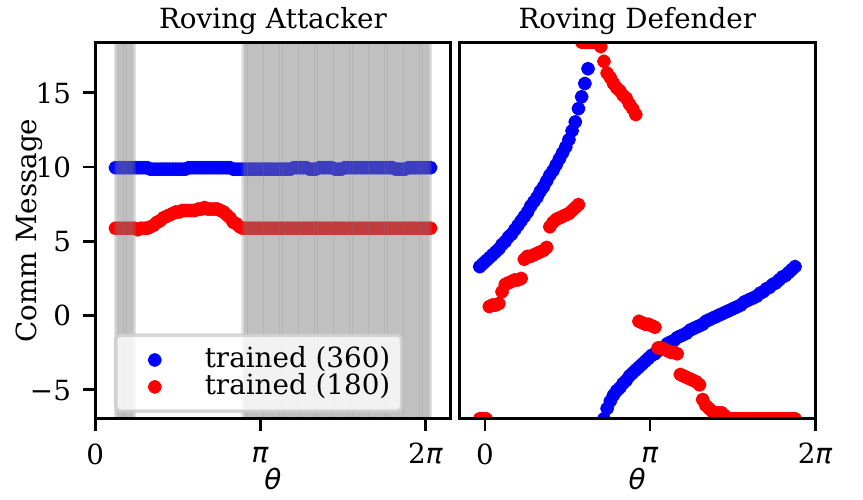}
\vspace{-7mm}
\caption{Impact of attacker (left) and defender (right) position on communication message. Grey are areas where attacker is not visible to 180 defender.}
\label{fig:com_visualization}
\vspace{-5mm}
\end{figure}

\fig\ref{fig:com_visualization} provides qualitative insight into how these communication channels are being used in a 7~vs.~3 game with a message width of one.
The left plot shows how one defender's broadcast value changes as one intruder is moved around the playing field.
The right plot shows how that defender's broadcast value changes as its own position is varied.
As expected, the \SI{360}{\degree} field of view policy generates messages that are a function only of the defender's own position, and not attacker locations.
The agent has learned to omit redundant information (attackers seen by everyone else) and summarize critical information (the self x,y coordinates encoded as an angular value).
Agents with only a \SI{180}{\degree} field of view must learn a more sophisticated policy.
Their message values are a function of the defender position, but observed attackers are also taken into account.

\section{DISCUSSION}

A motivation for this work was the recurring observation that due to the distributed nature of neural network computation, many neural network mechanisms (\textit{i.e.} \textit{PIN}, quantization) have clear application to multi-agent systems. For example, we may want to impose other invariances, such as spatial invariance or a graph connectivity structure through \textit{Convolutional Neural Networks}~\cite{goodfellow_deep_2016} and their generalization, the \textit{Graph Neural Network}~\cite{gama_convolutional_2019}. In addition, \textit{Fully Convolutional Networks}~\cite{long_fully_2015} could address variable size inputs.

Originally developed to improve network training, \textit{batch normalization}~\cite{ioffe_batch_2015} can standardize the range of the message activations, which increases comparability of the quantization process across models. In addition, we can apply $L_{1}$, $L_{2}$ regularization to the communication message. These regularization techniques are justified by the Minimum Description Length (MDL) principle~\cite{barron_minimum_1991}, which finds the best compression of data, or in our case, message. Dropout~\cite{srivastava_dropout_2014} was originally developed to prevent overfitting by randomly dropping connections during training. In real world multi-agent scenarios, communications from agents may be dropped, and we can model this phenomenon during training by using a dropout layer at the team communication level. When viewed in the lens of multi-agent systems, all of these mechanisms gain an additional, practical interpretation, strengthening our belief that neural networks should play a fundamental role in multi-agent systems.

\section{CONCLUSIONS}


This paper poses a policy decentralization problem for multirobot teams: given a centralized policy \(\Phi_c\) which is a function of the full joint state, find an equivalent decentalized policy \(\Phi_d\) which can be implemented by agents obeying local observation and communications constraints.
We suggest a construction of \(\Phi_d\) in terms of individual agent communication functions \(\psi_i\) and action functions \(\mu_i\), and we describe sufficient conditions under which such a \(\Phi_d\) will exist.

We describe a modular neural network architecture for training teams of identical agents which naturally addresses variable team size and obeys certain invariants inherent to multirobot policies.
This architecture provides a machinery for automatically approximating \(\psi_i\) and \(\mu_i\), providing both communication and action policies at the agent level.

We evaluate our approach in a multi-agent perimeter defense game.
Simulation results demonstrate that compact communication policies are learned, and that at test time these agents can generalize to teams larger than those they were trained in.

A wide range of frameworks and analytic tools have be effectively applied to centralized multiagent problems.
Learning to leverage this past success to develop distributed, communication aware policies is a path to more scalable, resilient, and effective robot teams.

\addtolength{\textheight}{-7.4cm}   

\bibliographystyle{IEEEtran}
\bibliography{IEEEabrv,references}

\begin{thebibliography}{10}
\providecommand{\url}[1]{#1}
\csname url@samestyle\endcsname
\providecommand{\newblock}{\relax}
\providecommand{\bibinfo}[2]{#2}
\providecommand{\BIBentrySTDinterwordspacing}{\spaceskip=0pt\relax}
\providecommand{\BIBentryALTinterwordstretchfactor}{4}
\providecommand{\BIBentryALTinterwordspacing}{\spaceskip=\fontdimen2\font plus
\BIBentryALTinterwordstretchfactor\fontdimen3\font minus
  \fontdimen4\font\relax}
\providecommand{\BIBforeignlanguage}[2]{{%
\expandafter\ifx\csname l@#1\endcsname\relax
\typeout{** WARNING: IEEEtran.bst: No hyphenation pattern has been}%
\typeout{** loaded for the language `#1'. Using the pattern for}%
\typeout{** the default language instead.}%
\else
\language=\csname l@#1\endcsname
\fi
#2}}
\providecommand{\BIBdecl}{\relax}
\BIBdecl

\bibitem{saeedi_multiplerobot_2016}
S.~Saeedi, M.~Trentini, M.~Seto, and H.~Li, ``Multiple-robot simultaneous
  localization and mapping: A review,'' \emph{Journal of Field Robotics},
  vol.~33, no.~1, pp. 3--46, 2016.

\bibitem{howard_multirobot_2006}
A.~Howard, ``Multi-robot simultaneous localization and mapping using particle
  filters,'' \emph{The International Journal of Robotics Research}, vol.~25,
  no.~12, pp. 1243--1256, Dec. 2006.

\bibitem{liu_multirobot_2016}
Y.~Liu and G.~Nejat, ``Multirobot cooperative learning for semiautonomous
  control in urban search and rescue applications,'' \emph{Journal of Field
  Robotics}, vol.~33, no.~4, pp. 512--536, 2016.

\bibitem{kantor_distributed_2006}
G.~Kantor, S.~Singh, R.~Peterson, D.~Rus, A.~Das, V.~Kumar, G.~Pereira, and
  J.~Spletzer, ``Distributed search and rescue with robot and sensor teams,''
  in \emph{Field and {{Service Robotics}}}, ser. Springer {{Tracts}} in
  {{Advanced Robotics}}.\hskip 1em plus 0.5em minus 0.4em\relax {Springer,
  Berlin, Heidelberg}, 2006, vol.~24, pp. 529--538.

\bibitem{barrientos_aerial_2011}
A.~Barrientos, J.~Colorado, J.~del Cerro, A.~Martinez, C.~Rossi, D.~Sanz, and
  J.~a. Valente, ``Aerial remote sensing in agriculture: {{A}} practical
  approach to area coverage and path planning for fleets of mini aerial
  robots,'' \emph{Journal of Field Robotics}, vol.~28, no.~5, pp. 667--689,
  2011.

\bibitem{kazmi_adaptive_2011}
W.~Kazmi, M.~Bisgaard, F.~{Garcia-Ruiz}, K.~D. Hansen, and A.~la~{Cour-Harbo},
  ``Adaptive surveying and early treatment of crops with a team of autonomous
  vehicles,'' in \emph{Proceedings of the 5th {{European Conference}} on
  {{Mobile Robots}} ({{ECMR}} 2011)}, Orebro, Sweden, 2011, pp. 253--258.

\bibitem{dandrea_revolution_2012}
R.~D'Andrea, ``A revolution in the warehouse: A retrospective on {{Kiva
  Systems}} and the grand challenges ahead,'' \emph{IEEE Transactions on
  Automation Science and Engineering}, vol.~9, no.~4, pp. 638--639, Oct. 2012.

\bibitem{miikkulainen_multiagent_2012}
R.~Miikkulainen, E.~Feasley, L.~Johnson, I.~Karpov, P.~Rajagopalan, A.~Rawal,
  and W.~Tansey, ``Multiagent learning through neuroevolution,'' in
  \emph{Advances in {{Computational Intelligence}}}.\hskip 1em plus 0.5em minus
  0.4em\relax {Springer, Berlin, Heidelberg}, Jun. 2012, pp. 24--46.

\bibitem{busoniu_comprehensive_2008}
L.~Busoniu, R.~Babuska, and B.~De~Schutter, ``A comprehensive survey of
  multiagent reinforcement learning,'' \emph{IEEE Transactions on Systems, Man,
  and Cybernetics, Part C: Applications and Reviews}, vol.~38, no.~2, pp.
  156--172, Mar. 2008.

\bibitem{kraemer_multiagent_2016}
L.~Kraemer and B.~Banerjee, ``Multi-agent reinforcement learning as a rehearsal
  for decentralized planning,'' \emph{Neurocomputing}, vol. 190, pp. 82--94,
  May 2016.

\bibitem{oliehoek_optimal_2008}
F.~A. Oliehoek, M.~T.~J. Spaan, and N.~Vlassis, ``Optimal and approximate
  {{Q}}-value functions for decentralized {{POMDPs}},'' \emph{Journal of
  Artificial Intelligence Research}, vol.~32, pp. 289--353, May 2008.

\bibitem{lowe_multiagent_2017}
R.~Lowe, Y.~Wu, A.~Tamar, J.~Harb, P.~Abbeel, and I.~Mordatch, ``Multi-agent
  actor-critic for mixed cooperative-competitive environments,'' in
  \emph{Advances in {{Neural Information Processing Systems}} 30 ({{NIPS}}
  2017)}, 2017, pp. 6379--6390.

\bibitem{foerster_counterfactual_2018}
J.~N. Foerster, G.~Farquhar, T.~Afouras, N.~Nardelli, and S.~Whiteson,
  ``Counterfactual multi-agent policy gradients,'' in \emph{Thirty-{{Second
  AAAI Conference}} on {{Artificial Intelligence}}}, Apr. 2018.

\bibitem{foerster_learning_2016}
J.~N. Foerster, Y.~M. Assael, N.~{de Freitas}, and S.~Whiteson, ``Learning to
  communicate to solve riddles with deep distributed recurrent
  {{Q}}-{{Networks}},'' \emph{preprint, arXiv:1602.02672}, Feb. 2016.

\bibitem{sukhbaatar_learning_2016}
S.~Sukhbaatar, A.~Szlam, and R.~Fergus, ``Learning multiagent communication
  with backpropagation,'' in \emph{Advances in {{Neural Information Processing
  Systems}} 29 ({{NIPS}} 2016)}, 2016, pp. 2252--2260.

\bibitem{dobbe_fully_2017}
R.~Dobbe, D.~{Fridovich-Keil}, and C.~Tomlin, ``Fully decentralized policies
  for multi-agent systems: An information theoretic approach,'' in
  \emph{Advances in {{Neural Information Processing Systems}} 30 ({{NIPS}}
  2017)}, 2017, pp. 2941--2950.

\bibitem{boutilier_planning_1996}
C.~Boutilier, ``Planning, learning and coordination in multiagent decision
  processes,'' in \emph{Proceedings of the 6th Conference on {{Theoretical}}
  Aspects of Rationality and Knowledge}, Mar. 1996, pp. 195--210.

\bibitem{amato_decentralized_2013}
C.~Amato, G.~Chowdhary, A.~Geramifard, N.~K. Ure, and M.~J. Kochenderfer,
  ``Decentralized control of partially observable {{Markov}} decision
  processes,'' in \emph{52nd {{IEEE Conference}} on {{Decision}} and
  {{Control}}}, Firenze, Dec. 2013, pp. 2398--2405.

\bibitem{zaheer_deep_2017}
M.~Zaheer, S.~Kottur, S.~Ravanbakhsh, B.~Poczos, R.~R. Salakhutdinov, and A.~J.
  Smola, ``Deep sets,'' in \emph{Advances in {{Neural Information Processing
  Systems}} 30 ({{NIPS}} 2017)}, 2017, pp. 3391--3401.

\bibitem{ravanbakhsh_deep_2017}
S.~Ravanbakhsh, J.~Schneider, and B.~Poczos, ``Deep learning with sets and
  point clouds,'' \emph{preprint, arXiv:1611.04500}, Feb. 2017.

\bibitem{ravanbakhsh_equivariance_2017}
S.~Ravanbakhsh, J.~Schneider, and B.~P\'oczos, ``Equivariance through
  parameter-sharing,'' in \emph{Proceedings of the 34th {{International
  Conference}} on {{Machine Learning}}}, Sydney, Australia, Jun. 2017, pp.
  2892--2901.

\bibitem{goodfellow_deep_2016}
I.~Goodfellow, Y.~Bengio, and A.~Courville, \emph{Deep Learning}.\hskip 1em
  plus 0.5em minus 0.4em\relax {MIT Press}, Nov. 2016.

\bibitem{gupta_deep_2015}
S.~Gupta, A.~Agrawal, K.~Gopalakrishnan, and P.~Narayanan, ``Deep learning with
  limited numerical precision,'' in \emph{International {{Conference}} on
  {{Machine Learning}}}, Lille, France, Jun. 2015, pp. 1737--1746.

\bibitem{courbariaux_binaryconnect_2015}
M.~Courbariaux, Y.~Bengio, and J.-P. David, ``{{BinaryConnect}}: Training deep
  neural networks with binary weights during propagations,'' in \emph{Advances
  in {{Neural Information Processing Systems}} 28 ({{NIPS}} 2015)}, 2015, pp.
  3123--3131.

\bibitem{shishika_localgame_2018}
D.~Shishika and V.~Kumar, ``Local-game decomposition for multiplayer
  perimeter-defense problem,'' in \emph{2018 {{IEEE Conference}} on
  {{Decision}} and {{Control}} ({{CDC}})}, Miami Beach, Dec. 2018, pp.
  2093--2100.

\bibitem{gama_convolutional_2019}
F.~Gama, A.~G. Marques, G.~Leus, and A.~Ribeiro, ``Convolutional neural network
  architectures for signals supported on graphs,'' \emph{IEEE Transactions on
  Signal Processing}, vol.~67, no.~4, pp. 1034--1049, Feb. 2019.

\bibitem{long_fully_2015}
J.~Long, E.~Shelhamer, and T.~Darrell, ``Fully convolutional networks for
  semantic segmentation,'' in \emph{Proceedings of the {{IEEE Conference}} on
  {{Computer Vision}} and {{Pattern Recognition}}}, 2015, pp. 3431--3440.

\bibitem{ioffe_batch_2015}
S.~Ioffe and C.~Szegedy, ``Batch normalization: Accelerating deep network
  training by reducing internal covariate shift,'' in \emph{International
  {{Conference}} on {{Machine Learning}}}, Jun. 2015, pp. 448--456.

\bibitem{barron_minimum_1991}
A.~R. Barron and T.~M. Cover, ``Minimum complexity density estimation,''
  \emph{IEEE Transactions on Information Theory}, vol.~37, no.~4, pp.
  1034--1054, Jul. 1991.

\bibitem{srivastava_dropout_2014}
N.~Srivastava, G.~Hinton, A.~Krizhevsky, I.~Sutskever, and R.~Salakhutdinov,
  ``Dropout: A simple way to prevent neural networks from overfitting,''
  \emph{Journal of Machine Learning Research}, vol.~15, no.~1, pp. 1929--1958,
  Jan. 2014.

\end{thebibliography}

\end{document}